\documentclass[letterpaper, 10 pt, conference]{ieeeconf}  

\IEEEoverridecommandlockouts                              

\usepackage{graphicx}
\usepackage{amsmath} 

\title{\LARGE \bf Breaking Task Impasses Quickly: Adaptive Neuro-Symbolic Learning for Open-World Robotics
}

\author{Pierrick Lorang$^{1}$
\thanks{$^{1}$Pierrick Lorang is with the Human-Robot Interaction Lab,
        Tufts University, 419 Boston Ave, Medford, MA 02155, United-States
        {\tt\small pierrick.lorang@tufts.edu}}%
}

\begin{document}

\maketitle
\thispagestyle{empty}
\pagestyle{empty}


\begin{abstract}
Adapting to unforeseen novelties in open-world environments remains a major challenge for autonomous systems. While hybrid planning and reinforcement learning (RL) approaches show promise, they often suffer from sample inefficiency, slow adaptation, and catastrophic forgetting. We present a neuro-symbolic framework integrating hierarchical abstractions, task and motion planning (TAMP), and reinforcement learning to enable rapid adaptation in robotics. Our architecture combines symbolic goal-oriented learning and world model-based exploration to facilitate rapid adaptation to environmental changes. Validated in robotic manipulation and autonomous driving, our approach achieves faster convergence, improved sample efficiency, and superior robustness over state-of-the-art hybrid methods, demonstrating its potential for real-world deployment.
\end{abstract}

\section{Introduction}
As robots transition from controlled environments to dynamic, open-world settings, their ability to handle unforeseen changes becomes critical. Traditional RL methods, while powerful, suffer from sample inefficiency, slow adaptation~\cite{qu2020minimalistic, LESORT202052}, and catastrophic forgetting~\cite{kemker2018measuring}. Symbolic planning offers structured reasoning but lacks the flexibility to adapt to unpredictable variations.

Hybrid approaches combining symbolic planning with RL help mitigate these challenges by leveraging structured abstraction alongside learning-based adaptability. However, they still struggle with sample efficiency and adaptation speed, particularly in dynamic settings requiring rapid adjustments~\cite{muhammad2021novelty, sarathy2021spotter, Nayyar_Verma_Srivastava_2022}. A key challenge is linking symbolic and neural representations to improve adaptation, exploration, and generalization~\cite{Xu_Fekri_2022, lorangetal2024iros}.

We introduce a neuro-symbolic framework that strengthens the interaction between learning and planning, enhancing adaptation in dynamic environments. Our key contributions include:
(1) A hierarchical structure extending beyond bi-level systems, integrating nested controllers and action abstractions for adaptive learning and planning. (2)Symbolic goal-oriented learning, refining RL control through high-level task abstraction to improve generalization. (3) Curiosity-driven symbolic imagination, leveraging world models for more effective exploration and learning. (4) Object-oriented skill learning for problem simplification and cross-embodiment transfer.

\begin{figure}[t]
  \centering
  \begin{minipage}[t]{0.44\textwidth}
    \includegraphics[width=0.99\textwidth]{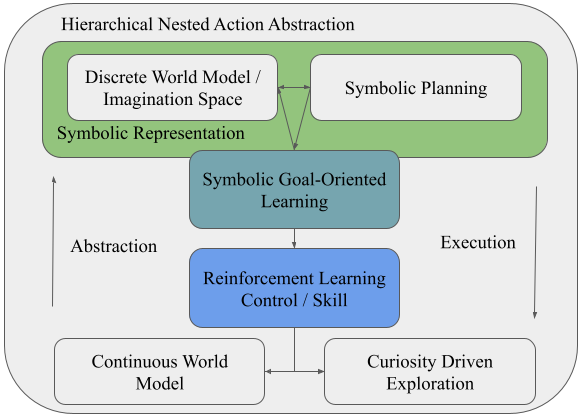}
    \caption{High-level architecture of our neuro-symbolic framework integrating symbolic planning, RL, and skill abstraction for adaptive learning.}
    \label{fig:hybrid_planning_learning}
  \end{minipage}
\end{figure}

\section{Related Work}
Recent work integrates symbolic planning with reinforcement learning (RL) to tackle open-world challenges. Hybrid approaches leverage symbolic reasoning and RL adaptability but often suffer from data inefficiency and slow training, especially in continuous settings~\cite{liu2023ai, goeletal24aij}. They also struggle with recovering from planning failures in rapidly changing environments~\cite{chen2022single, goel2022rapidlearn}.
Efforts to improve data efficiency include integrating goal-oriented RL~\cite{lorang2022speeding} and hierarchical frameworks~\cite{peorl-Yang, lorangetal24icra} into hybrid methodologies, though they often rely on sparse or human-aided rewards. Advances in symbolic world models~\cite{Arora_Fiorino_Pellier_Métivier_Pesty_2018, Konidaris_Kaelbling_Lozano-Perez_2018} and curiosity-driven exploration~\cite{curtis2020flexible, sartor2023intrinsically} aid adaptation and exploration. Reward densification via bi-level heuristics and temporal logic-based reward machines further address sparse feedback~\cite{Icarte2020RewardME, seo2020trajectorywise, Hu_Wang_Jia_Wang_Chen_Hao_Wu_Fan, Stadie_Zhang_Ba_2020}. Our work integrates these approaches to enhance adaptation and learning in open-world environments.

\section{Preliminaries}

Symbolic planning involves a domain described by a set of entities \( \mathcal{E} \), predicates \( \mathcal{F} \), symbolic states \( \mathcal{S} \), and action operators \( \mathcal{O} \), where each operator \( o_i \in \mathcal{O} \) has preconditions \( \psi_i \) and effects \( \omega_i \), both from \( \mathcal{F} \). A planning task is represented as \( T = \langle \mathcal{E}, \mathcal{F}, \mathcal{O}, s_0, s_g \rangle \), where \( s_0 \) is the initial state and \( s_g \) is the goal state, and its solution is a sequence of operators forming a plan \( \mathcal{P} \). In reinforcement learning (RL), an agent interacts with an environment modeled as an MDP \( M = \langle \mathcal{\widetilde{S}}, \mathcal{A}, R, \tau, \gamma \rangle \), where \(\mathcal{A}\) consists of the primitive actions and \(\mathcal{S}\) the sub-symbolic continuous space, aiming to learn a policy \( \pi(a|\tilde{s}) \) that maximizes cumulative reward. The Integrated Planning Task (IPT) framework combines planning and learning, where an IPT \( \mathcal{T} = \langle T, M, d, e \rangle \)~\cite{sarathy2021spotter} integrates a STRIPS task with an MDP, enabling adaptive learning through executors \( x = \langle I, \pi, \beta \rangle \). Here, \( I \) represents the initiation condition, \( \pi \) is the reinforcement learning policy, and \( \beta \) is the termination condition.

\section{Methodology}

\subsection{Hierarchical Nested Action Abstraction}

Our framework structures actions hierarchically, integrating symbolic planning, RL, and HRL~\cite{lorang2022speeding, lorangetal24icra}. Each operator \( o_i \) in a plan \(\mathcal{P}\) is associated with executors \(\mathcal{X}_i\), which can be either control executors operating in an MDP \( M^c = \langle \mathcal{\widetilde{S}}, \mathcal{A}, R_i, \tau, \gamma \rangle \) or skill executors acting in a skill-SMDP \( M^s = \langle \mathcal{\widetilde{S}}, \mathcal{O}, R_i, \tau, \gamma \rangle \), where \(\mathcal{O}\) consists of higher-level options. If all executors in \(\mathcal{X}_i\) fail, the agent determines whether the novelty is local or global. For local novelties, HRL refines or learns a new skill executor \( x_{\textit{new}}^s = \langle I_x, \pi^s_x, \beta_x \rangle \). For global novelties, a new control executor \( x_{\textit{new}}^c = \langle I_x, \pi^c_x, \beta_x \rangle \) is trained. This hierarchical structure reduces action space complexity and improves adaptability by leveraging existing behaviors in novel contexts.

\subsection{Symbolic Goal-Oriented Learning of Controls}

We use symbolic goal-oriented learning for control discovery~\cite{lorangetal2024iros}. Instead of defining goals in a high-dimensional state space, we define goal spaces within a symbolic abstraction \( \mathcal{G} \subseteq \mathcal{S} \), where \( \mathcal{S} \) is a human-interpretable space derived from \( \mathcal{\widetilde{S}} \). This abstraction improves learning efficiency by aligning goal achievement with task-level predicates. An executor \( x \) is defined as \( x = \langle I, \pi_{g}, \beta_{g} \rangle \), where \( I \) is the initiation function, \( \pi_{g} \) is the symbolic goal-conditioned policy, and \( \beta_{g} \) ensures goal satisfaction. We use hindsight experience replay (HER)~\cite{DBLP:journals/corr/AndrychowiczWRS17} within the symbolic framework to increase sample efficiency by reinterpretation of past experiences with different goals.

\subsection{Curiosity-Driven Imagination}

We integrate a curiosity-driven exploration module~\cite{pmlr-v70-pathak17a}, motivating the agent to explore underexplored states using an intrinsic reward function \( \mathcal{R}_{\text{intrinsic}} \). This drives the discovery of novel symbolic transitions in continuous state spaces, which are then abstracted into a symbolic world model. A symbolic lifted operator learner abstracts these transitions as PDDL operators in the imaginary space. This allows the agent to reason about potential symbolic actions, such as imagining picking an orange after learning to pick an apple. This abstraction enhances the agent's ability to plan and build reward machines across tasks, guiding its learning and improving its efficiency in dynamic environments.

\section{Experimental Validation}

We evaluate key components of our architecture through ablation studies. We test the ``PRM\&ICM Framework" and ``HyGOAL Framework" in RoboSuite~\cite{robosuite2020} with the ``Pick and Place Can" task, where a robotic arm places a can in a bin. The ``Hierarchy Framework" is tested in the CARLA~\cite{Dosovitskiy17} driving simulation for more complex abstraction scenarios. In both environments, we introduce two scenarios with five novelties each, where novelties obstruct essential symbolic operators. 

Agents were trained until meeting a convergence criterion: a success rate above $80\%$ or a maximum of $500,000$ steps in the Pick\&Place environment, and $2$ million steps in CARLA. An episode has a maximum of $1,000$ interactions. Results were averaged over $10$ seeds per agent, with performance evaluated every $20,000$ steps by running $20$ episodes and calculating the mean success rate. The same RL hyperparameters were applied across novelties. Key metrics were $\text{T}_{\text{adapt}}$ (time steps to convergence) and $\text{Success Rate}_{\text{post-training}}$ (success rate at convergence). Experimental details are available in~\cite{lorangetal24icra, lorangetal2024iros}.

\begin{table}[ht]
\centering
\scriptsize
\resizebox{\linewidth}{!}{%
\begin{tabular}{|c|c|c|c|c|c|}
\hline
\multicolumn{6}{|c|}{\textit{Results of the Curiosity-Driven Imagination in Pick\&Place}} \\
\hline
\textbf{Success Rate} & \textbf{Hole} & \textbf{Elevated} & \textbf{Obstacle} & \textbf{Door} & \textbf{Light-off} \\
\hline
PRM\&ICM & \textbf{0.76} & \textbf{1.00} & \textbf{0.71} & \textbf{0.92} & \textbf{0.80 }\\
Best Baseline & 0.38 & 0.88 & 0.65 & 0.89 & 0.77 \\
\hline
\textbf{$\text{T}_{\text{adapt}}$} & \textbf{Hole} & \textbf{Elevated} & \textbf{Obstacle} & \textbf{Door} & \textbf{Light-off} \\
\hline
PRM\&ICM & \textbf{290} & \textbf{98} & 385 & 210 & 290 \\
Best Baseline & 400 & 180 & \textbf{370} & \textbf{130} & \textbf{280} \\
\hline
\multicolumn{6}{|c|}{\textit{Results of Symbolic Goal-Oriented learning in Pick\&Place}} \\
\hline
\textbf{Success Rate} & \textbf{Hole} & \textbf{Elevated} & \textbf{Obstacle} & \textbf{Door} & \textbf{Light-off} \\
\hline
HyGOAL & \textbf{0.41} & \textbf{0.91} & 0.38 & \textbf{0.90} & \textbf{0.80} \\
Best Baseline & 0.21 & 0.84 & 0.66 & 0.72 & 0.73 \\
\hline
\textbf{$\text{T}_{\text{adapt}}$} & \textbf{Hole} & \textbf{Elevated} & \textbf{Obstacle} & \textbf{Door} & \textbf{Light-off} \\
\hline
HyGOAL & \textbf{398} & \textbf{166} & 390 & \textbf{138} & \textbf{264} \\
Best Baseline & 416 & 214 & 362 & 294 & 310 \\

\hline
\multicolumn{6}{|c|}{\textit{Results of the Nested Hierarchy in Autonomous Driving Domain}} \\
\hline
\textbf{Success Rate} & \textbf{Ice} & \textbf{Tire\&Rain} & \textbf{Obstacle} & \textbf{Obst\&Ice} & \textbf{Obst\&Mist} \\
\hline
Hierarchy & \textbf{0.89} & \textbf{0.9} & \textbf{0.72} & \textbf{0.6} & 0.81 \\
Best Baseline & \textbf{0.89} & \textbf{0.9} & 0.68 & 0.2 & \textbf{0.83} \\
\hline
\textbf{$\text{T}_{\text{adapt}}$} & \textbf{Ice} & \textbf{Tire\&Rain} & \textbf{Obstacle} & \textbf{Obst\&Ice} & \textbf{Obst\&Mist} \\
\hline
Hierarchy & \textbf{2000} & \textbf{2000} & \textbf{56} & \textbf{50} & \textbf{37} \\
Best Baseline & \textbf{2000} & \textbf{2000} & 700 & 700 & 700 \\
\hline
\end{tabular}
}
\label{tab:results}
\end{table}

The ablation studies (preliminary results in Table 1) show that individual components of our architecture outperform the baselines in $13$ out of $15$ scenarios, with an average improvement of $20\%$, and achieve faster convergence in $11$ cases, reducing training time by $54\%$ on average. These results highlight the potential of integrating these components into a unified architecture for adaptive neuro-symbolic learning in open-world robotics.

\section{Conclusion}
We introduced a neuro-symbolic framework that combines hierarchical action abstraction, symbolic goal-oriented learning, and curiosity-driven exploration to improve adaptation in open-world robotics. By tightly integrating symbolic reasoning with reinforcement learning, our approach enhances sample efficiency and accelerates adaptation. Experimental results in robotic manipulation and autonomous driving demonstrate faster convergence and superior performance of individual components compared to state-of-the-art hybrid methods, suggesting that a unified architecture can further boost adaptive neuro-symbolic learning. These findings underscore the potential of our architecture for real-world deployment. Future work will explore neuro-symbolic imitation learning to incorporate efficient human feedback and demonstrations, further enabling rapid adaptation in dynamic environments.

\bibliographystyle{IEEEtran}
\bibliography{main}

\end{document}